\documentclass[preprint,12pt]{elsarticle}

\usepackage{graphicx}
\usepackage{wrapfig}

\usepackage{comment}
\usepackage{arydshln}

\usepackage{amsthm}

\newtheorem{theorem}{Theorem}

\theoremstyle{definition}

\usepackage[utf8]{inputenc} 
\usepackage[T1]{fontenc}    
\usepackage{hyperref}       
\usepackage{url}            
\usepackage{booktabs}       
\usepackage{amsfonts}       
\usepackage{amsmath,amssymb}        
\usepackage{nicefrac}       
\usepackage{microtype}      
\usepackage{xcolor}         
\usepackage{multirow}

\usepackage{xspace}

\newcommand{\Method}{RSPose\xspace}

\journal{Neurocomputing}

\begin{document}

\begin{frontmatter}



\title{RSPose: Ranking Based Losses for Human Pose Estimation} 


\author[1]{Muhammed Can Keles}
\author[1]{Bedrettin Cetinkaya}
\author[1,2]{Sinan Kalkan}
\author[1,2]{Emre Akbas}

\affiliation[1]{organization={Department of Computer Engineering, Middle East Technical University},
            addressline={Üniversiteler Mahallesi, Dumlupınar Bulvarı No:1}, 
            city={Ankara},
            postcode={06800}, 
            country={Turkey}}

\affiliation[2]{organization={ROMER, Middle East Technical University},
            addressline={Üniversiteler Mahallesi, Dumlupınar Bulvarı No:1}, 
            city={Ankara},
            postcode={06800}, 
            country={Turkey}}

\begin{abstract}
While heatmap-based human pose estimation methods
have shown strong performance, they suffer from
three main  problems: 
(P1) Commonly used ``Mean Squared Error (MSE) Loss'' may not always improve joint localization because it penalizes all pixel deviations equally, without focusing explicitly on sharpening and correctly localizing the peak corresponding to the joint; 
(P2) heatmaps are spatially and class-wise imbalanced; and, 
(P3) there is a discrepancy between the evaluation metric (i.e., mAP) and the loss functions. 
We propose ranking-based losses to address these issues.  
Both theoretically and empirically, we show that our proposed losses are superior to commonly used heatmap losses (MSE, KL-Divergence). 
Our losses considerably increase the correlation between confidence scores and localization qualities, which is desirable because higher correlation leads to more accurate instance selection during Non-Maximum Suppression (NMS) and better Average Precision (mAP) performance.  We refer to the models trained with our losses as \Method.
We show the effectiveness of \Method across two different modes: one-dimensional and two-dimensional heatmaps,  on three different datasets (COCO, CrowdPose, MPII). 
To the best of our knowledge, we are the first to propose losses that align with the evaluation metric (mAP) for human pose estimation. 
\Method outperforms the previous state of the art on the COCO-val set and achieves an mAP score of 79.9 with ViTPose-H, a vision transformer model for human pose estimation. 
We  also improve SimCC Resnet-50, a coordinate classification-based pose estimation method, by 1.5 AP on the COCO-val set, achieving 73.6 AP.
\end{abstract}


\end{frontmatter}



\section{Introduction}

\begin{figure}
    \centering
        \includegraphics[width=\linewidth]{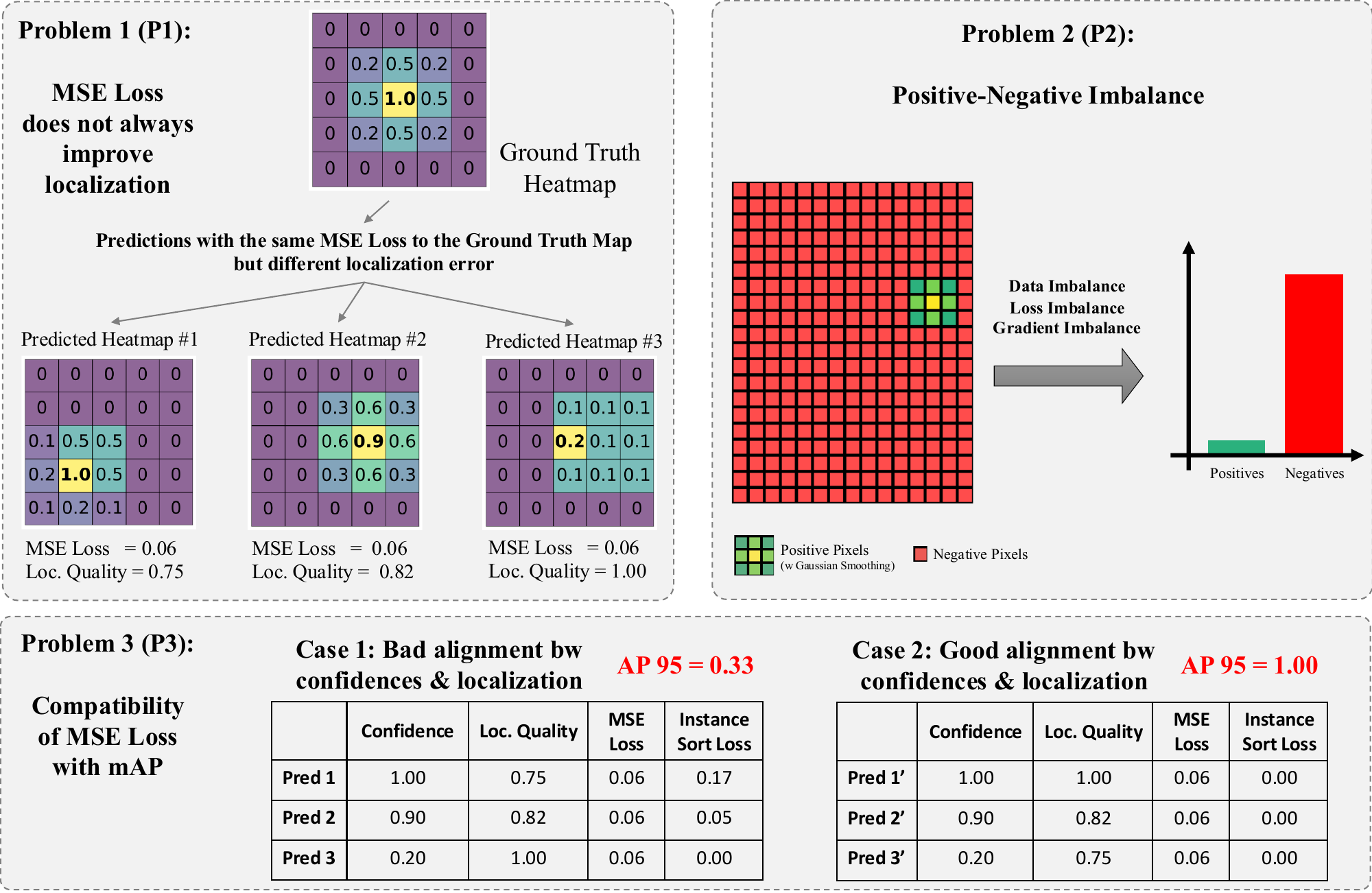}
    \caption{We highlight 3 problems when training heatmap based human pose estimation models with MSE. \textbf{(P1)} MSE loss does not always enhance localization, but causes models to learn the non-GT pixels better instead of learning optimal localization.
    \textbf{(P2)} Heatmaps are very sparse and imbalanced. MSE Loss does not provide balanced gradients for positive  and negative pixels.
    \textbf{(P3)} MSE Loss does not necessarily optimize the ranking alignment between localization qualities and confidences, which is crucial for the evaluation measure mAP.
    See Section \ref{sect:how_we_address_problems} for a detailed discussion on how our methods address these problems.
    }
    \label{fig:problems}
\end{figure}

Heatmap-based methods for human pose estimation have  been widely adopted  due to their strong performance \cite{NEURIPS2022_fbb10d31, Geng_2023_CVPR, Lu_2024_CVPR,Liu_2023_ICCV}.
These methods predict a heatmap for each joint, where each pixel encodes a pseudo-probability indicating the likelihood of the joint being located at that location. 
Heatmap-based methods have shown better performance compared to regression-based methods.
Top-down human pose estimation methods are typically trained by  minimizing the Mean-Squared Error (MSE) Loss between the predicted and ground truth heatmaps, where the ground truth heatmap is generated as a 2D Gaussian centered at the annotated joint location. 
Despite their success, these methods suffer from three key problems (P1-P3):  

\noindent\textbf{(P1) MSE Loss may not always improve localization quality.} MSE Loss minimizes pixel-wise differences across the entire heatmap, which can lead the model to focus on regions that have little influence on joint localization accuracy (Figure \ref{fig:problems}).

\noindent\textbf{(P2) Heatmaps incur positive-negative imbalance.} Heatmaps used in pose estimation are {highly imbalanced}, containing a single positive pixel surrounded by a large number of negative ones (as shown in Figure \ref{fig:problems} and analyzed in Section \ref{sect:analysis}). For instance, in a common 64×48 heatmap, the positive-to-negative (P/N) ratio is 1:3072. Unless handled, this  leads to severe imbalance in loss values as well as the gradients, which considerably affects model performance (analyzed in Section \ref{sect:analysis}). To address this, Gaussian label smoothing combined with MSE loss is commonly employed. While this approach yields good results, it also introduces limitations. 

\noindent\textbf{(P3) Discrepancy between  the evaluation measure and the training objective.} The commonly used evaluation measure, mAP, measures the alignment of instance localization qualities and instance confidences. While MSE supervises the model to learn reasonable confidence scores, it does not directly optimize mAP (Figure \ref{fig:problems}).

\textbf{To address these problems}, we propose ranking-based loss functions, inspired by their success in object detection and instance segmentation tasks \cite{chen2020ap, oksuz2020ranking, Oksuz_2021_ICCV, ramzi2025optimization, pu2023rank, tang2022ranking,Cetinkaya_2024_CVPR}, where significant imbalance issues commonly arise. 
Specifically, we introduce \textbf{Spatial-RS Loss}, which trains the model to rank the ground truth logit or pixel above background logits or pixels, which benefits from the robustness of ranking-based approaches to imbalances in data \cite{oksuz2020ranking} (\textbf{addressing P2}). Furthermore, we propose \textbf{Instance-Sort Loss}, which enforces alignment between keypoint confidence scores and their corresponding localization quality (\textbf{addressing P1 and P3}).  We refer to the models trained with our losses as \Method.

We validate our approach across both 1D and 2D heatmap-based methods: SimCC-ResNet50 and SimCC-HRNet48 for 1D, and ViTPose-B, ViTPose-H, and HRNet-32 for 2D. Experiments are conducted on three standard benchmarks: COCO, CrowdPose, and MPII.
Our method demonstrates strong empirical performance. On the COCO-val set, we achieve 79.9 AP with ViTPose-H, surpassing the previous state of the art. Additionally, our losses significantly increase the correlation between predicted confidence scores and localization quality ($16+$ percentage points), which is critical for reliable NMS and improved mAP.

Our contributions in this paper are threefold: \textbf{(1)} We are the first to successfully adapt and apply ranking based losses for human pose estimation.
\textbf{(2)} Our ranking-based losses, namely Spatial-RS Loss and Instance-Sort Loss, consistently outperform standard loss
functions such as mean-squared-error and KL divergence on a diverse set of models and dataset.   
\textbf{(3)} We introduce Instance Sort Loss (Instance-Sort), a ranking-based objective that encourages stronger alignment between predicted confidence scores and localization accuracy. To our knowledge, this is the first loss in human pose estimation explicitly designed to be aligned with the mAP evaluation measure. Instance-Sort improves confidence-localization correlation and contributes to measurable performance gains.

\section{Related Work}
\label{gen_inst}

\textbf{Human Pose Estimation.} Human pose estimation has shown great progress recently with several different paradigms: top-down \cite{NEURIPS2022_fbb10d31, Geng_2023_CVPR, Lu_2024_CVPR, li2022simcc}, bottom-up \cite{geng2021bottom,luo2021rethinking,xue2022learning} and end-to-end \cite{liu2023group,yangexplicit}. Bottom-up methods and end-to-end methods are faster while providing inferior performance. Heatmap based methods have shown superior performance compared to regression based ones, in general. 

Commonly, heatmap based methods are trained with MSE Loss that is computed between the ground truth heatmap and the predicted heatmap. Previous work \cite{qu2022heatmap} has formulated heatmap prediction as a distribution matching problem, proposing the usage of Earth Movers Distance as a heatmap loss.

\textbf{Ranking based losses for computer vision. } Ranking based losses have shown strong performance in several computer vision problems such as object detection, instance segmentation, and edge detection. \cite{chen2020ap,Oksuz_2021_ICCV,Cetinkaya_2024_CVPR,yavuz2024bucketed}. Ranking based losses deal with both imbalance and uncertainty, as demonstrated in \cite{oksuz2020ranking,Oksuz_2021_ICCV,Cetinkaya_2024_CVPR}.

\textbf{Comparative Summary.} In this work, we propose an alternative loss to MSE for better handling the (i) imbalance, (ii) non-optimal localization, (iii) discrepancy between loss and evaluation metric problems for human pose estimation. To the best of our knowledge, we are the first to propose ranking based losses for human pose estimation and the first to study the loss-to-evaluation metric compatibility for human pose estimation.

\section{Pose Estimation: Preliminaries}
\label{sect:preliminaries}

\paragraph{Pose Estimation Problem Definition.} Given an image $I \in \mathbb{R}^{H\times W\times 3}$, human pose estimation is the problem of estimating $N$ keypoints $\mathcal{K}=\{k_1, ..., k_N\}$ where each keypoint $k_i$ corresponds to a spatial coordinate in $I$. We use a deep network $f(\cdot; \theta)$ with parameters $\theta$ to obtain a heatmap volume $\hat{\mathcal{H}}\in \mathbb{R}^{H\times W\times N}$ which includes one heatmap channel for each keypoint: $\hat{\mathcal{H}} = f(I; \theta)$. We denote the ground truth heatmap with $\mathcal{H} \in \mathbb{R}^{H\times W\times N}$. Although each type of keypoint has a single true-positive pixel, the ground-truth heatmap is typically generated by applying a 2D Gaussian kernel to smooth the label.
One dimensional heatmaps are also used.  SimCC \cite{li2022simcc} employs a combination of two independent 1D heatmaps -- one for the horizontal (x-axis) and one for the vertical (y-axis) coordinates. To achieve sub-pixel localization precision and mitigate quantization errors inherent in traditional heatmap-based methods, this approach discretizes each axis into $N_x = W \cdot k$ and $N_y = H \cdot k$ bins, where $W$ and $H$ are the image width and height, and $k$ is a splitting factor greater than or equal to 1. The ground truth for each keypoint is represented as a 1D Gaussian distribution centered at the true coordinate, providing a soft label that reflects spatial uncertainty. The model outputs predicted probability distributions $\hat{\mathbf{p}}_x$ and $\hat{\mathbf{p}}_y$ over these discretized bins: $(\hat{\mathbf{p}}_x, \hat{\mathbf{p}}_y) = f(I; \theta)$. We denote the ground truth distributions with $({\mathbf{p}}_x, {\mathbf{p}}_y)$.

\paragraph{Loss Functions for Pose Estimation.} In this paper, we limit our scope to heatmap-based pose estimation methods. 
The literature has adopted different loss functions for supervising a deep network's heatmap prediction $\hat{\mathcal{H}}$. For example, mean-squared error (MSE) loss is commonly used to penalize pixel-wise predictions as \cite{NEURIPS2022_fbb10d31, wang2020deep}:
\begin{equation}\label{eq:mse_loss}\footnotesize
\mathcal{L}_{\text{MSE}}(\hat{\mathcal{H}}, \mathcal{H}) = \frac{1}{WH} \sum_{x=1}^{W} \sum_{y=1}^{H} \left(\hat{\mathcal{H}}(x, y) - \mathcal{H}(x, y)\right)^2.
\end{equation}
Studies using the formulation proposed by SimCC \cite{li2022simcc} employ the Kullback-Leibler (KL) divergence to quantify the error between the predicted and ground truth distributions:

\begin{equation}\footnotesize
\log \mathbf{p}_x = \log\left(\mathrm{softmax}(\mathbf{z}_x \cdot \beta)\right), \;\;\;\;\;\;
\log \mathbf{p}_y = \log\left(\mathrm{softmax}(\mathbf{z}_y \cdot \beta)\right),
\end{equation}
\begin{equation}\footnotesize
\mathcal{L}_{\text{KL}} = \frac{1}{N_x} \sum_{i=1}^{N_x} D_\mathrm{KL}\left(\log \mathbf{p}_x^{(i)} \,\|\, \mathbf{y}_x^{(i)}\right) + \frac{1}{N_y} \sum_{j=1}^{N_y} D_\mathrm{KL}\left(\log \mathbf{p}_y^{(j)} \,\|\, \mathbf{y}_y^{(j)}\right).
\end{equation}

\section{Pose Estimation and the Problem of Imbalance}
\label{sect:analysis}

\begin{figure} 
  \centering
  \includegraphics[width=0.50\textwidth]{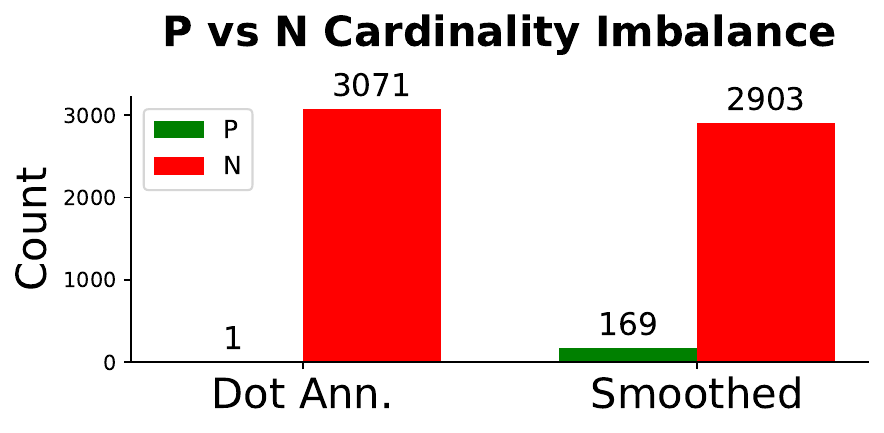}
  \caption{Positive-negative imbalance between positive and negative samples in heatmaps, with an input size of 256×196 and a heatmap size of 64×48. Left: One pixel is positive and other pixels are negative for a joint. Right: A region of pixels around a positive pixel are annotated a degree of positiveness using a Gaussian function centered at the positive pixel.}
  \label{fig:imbal_hist}
\end{figure}

\paragraph{Imbalance in Cardinalities.} Visual recognition problems often exhibit imbalance between positive ($\mathcal{P}$) and negative ($\mathcal{N}$) samples \cite{yang2022survey,saini2023tackling,oksuz2020imbalance,das2022supervised}. 
Human pose estimation also exhibits imbalance between positive and negative samples: For a joint, there is only a single positive pixel while the rest of the pixels are all negative. This so-called dot annotation is a case of severe imbalance (Figure \ref{fig:imbal_hist}), which can yield sub-optimal performances for the minority class.

This imbalance is often addressed in the literature by applying Gaussian label smoothing at the positive pixels. This way, a region of pixels around a positive pixel is annotated with a degree of positiveness based on their distance to the positive pixel. As illustrated in Figure \ref{fig:imbal_hist}, this produces more positive labels to balance training in favor of positive pixels.

\textbf{Imbalance in Loss and its Gradients.} To better see how the cardinality imbalance between positive ($\mathcal{P}$) and negative ($\mathcal{N}$) samples might affect the training of $f(\cdot; \theta)$, we can rewrite the MSE loss from Eq. \ref{eq:mse_loss} as:
\begin{equation}\footnotesize
\mathcal{L}_{\text{MSE}}(\hat{\mathcal{H}}, \mathcal{H}) = \frac{1}{WH} \left( \sum_{(x, y) \in \mathcal{P}} \ell_{xy} + \sum_{(x, y) \in \mathcal{N}} \ell_{xy} \right),
\end{equation}
where $\ell_{xy} = \left(\hat{\mathcal{H}}(x, y) - {\mathcal{H}}(x, y)\right)^2$. Since $|\mathcal{P}| \ll |\mathcal{N}|$, the total loss is dominated by background (negative) pixels, even though these pixels contribute little semantic value for keypoint localization. 

\begin{figure} 
    \includegraphics[width=0.99\linewidth]{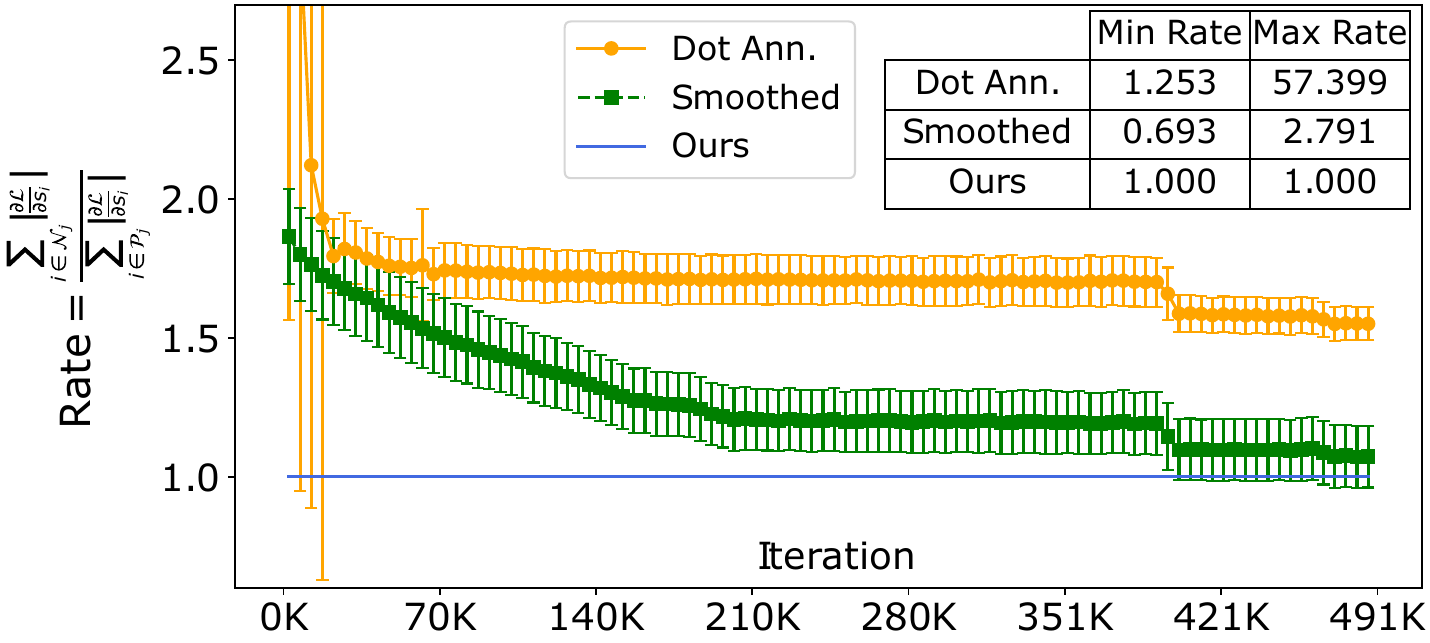}
    \caption{Cardinality imbalance between positive and negative keypoint samples causes imbalance between the gradients of positive and negative keypoint samples. Our proposed method (blue line) addressed this issue, providing a perfect balance between the gradients.}
    \label{fig:imbal_grad}
\end{figure}

The imbalance in the loss values, in turn, causes gradient descent to an imbalanced distribution of gradients between positives and negatives, as illustrated in Figure \ref{fig:imbal_grad}.

This imbalance between gradients can hinder optimization, as the model is implicitly incentivized to minimize prediction error over the many non-informative background pixels. To illustrate this, in Table \ref{tab:imbalance}, we investigate the effect of using Gaussian label smoothing and show that increasing the number of positive samples provides significant improvements.

\begin{table*}
    \centering
    \label{tab:imbalance}\footnotesize
    \begin{tabular}{lccccc}
        \toprule
        Method & Loss &  Label Type & Imbalance Ratio ($|\mathcal{N}|/|\mathcal{P}|$) & AP & AR \\
        \midrule
         ViTPose \cite{NEURIPS2022_fbb10d31} & MSE & Dot Annotation & 3071 &  67.7 & 77.3 \\ 
         ViTPose \cite{NEURIPS2022_fbb10d31} & MSE & Gaussian Label S. & 17 & 75.8 & 81.1 \\ 
         \cdashline{1-6}
         SimCC \cite{li2022simcc} & KL-Div. & Dot Annotation & along x: 383, along y: 511 & 71.3 & 77.3 \\
         SimCC \cite{li2022simcc} & KL-Div. & Gaussian Label S. & along x: 1.2, along y: 2 & 73.4 & 80.6 \\
         \bottomrule
    \end{tabular}
    \caption{The impact of cardinality imbalance on pose estimation performance. Increasing the number of positive samples by using Gaussian label smoothing improves the pose estimation performance. ViTPose uses a ViT-B backbone and SimCC uses a ResNet-50 backbone. (SB: SimpleBaseline \cite{xiao2018simple}).}
\end{table*}

\section{Method}
\label{sect:method}

In this section, we describe our  novel loss functions for human pose estimation: Spatial Rank \& Sort (RS) Loss and Instance Sort Loss. Both of these losses supervise the confidence scores of pixels. 

\textbf{Spatial RS Loss} aims to improve the localization quality of keypoints. It has two components: Spatial Rank Loss and Spatial Sort Loss. \textbf{Spatial Rank Loss} enforces the confidences of positive pixels to be higher than that of negative pixels. \textbf{Spatial Sort Loss} aims to sort positive pixels among themselves respect to their ground-truth labels. That is, a positive pixel with  a label, say $0.8$, is enforced to have a higher confidence score than another positive pixel with a lower value label. 

\textbf{Instance Sort Loss}, on the other hand, aims to adjust confidence scores across instances.  For a given joint type, it selects the highest-confidence  pixel from each person instance in the batch and sorts them by ``keypoint similarity'' (KS) score, which measures localization quality (and acts like an intersection over union for keypoints).

\subsection{Spatial Rank \& Sort Loss}

Our first contribution, Spatial Rank \& Sort Loss, has two components, Ranking Loss ($\mathcal{L}_{\text{S-Rank}}$) and Sorting Loss ($\mathcal{L}_{\text{S-Sort}}$).

\noindent\textbf{(1) Spatial Rank Loss} ($\mathcal{L}_{\text{S-Rank}}$) aims to rank positively labeled pixels  above negatively labeled ones (adapting the definitions in Oksuz et al. \cite{Oksuz_2021_ICCV}):
\begin{equation}\label{eq:l_rank}\footnotesize
\mathcal{L}_{\text{S-Rank}} = \frac{1}{|\mathcal{P}|} \sum_{i \in \mathcal{P}} ({\ell}_{\text{S-Rank}}(i) - \ell^*_{\text{S-Rank}}(i)),
\end{equation}
where ${\ell}_{\text{S-Rank}}(i)$ is the current ranking error for positive pixel $i$ whereas $\ell^*_{\text{S-Rank}}$ is the target ranking error. ${\ell}_{\text{S-Rank}}(i)$ measures precision for the $i^\text{th}$ positive pixel as the ratio of negative pixels having higher confidence than the positive pixel $i$:
\begin{equation}\footnotesize
    {\ell}_{\text{S-Rank}}(i) = \frac{\text{\# of false positives with }\hat{\mathbf{p}} > \hat{\mathbf{p}}_i}{\text{\# of pixels with }\hat{\mathbf{p}} > \hat{\mathbf{p}}_i } = \frac{N_\text{FP}(i)}{\text{rank}(i)} = \frac{\sum_{j\in \mathcal{N}} H(x_{ij})}{\sum_{j\in \mathcal{P}\cup\mathcal{N}} H(x_{ij})},
\end{equation}
with $x_{ij}=\hat{\mathbf{p}}_j - \hat{\mathbf{p}}_i$ signifying relative ordering between two pixels and $H(\cdot)$ is the step function: $H(x_{ij})= 1$ if $x_{ij}>0$ (i.e., $\hat{\mathbf{p}}_j > \hat{\mathbf{p}}_i$) and $0$ otherwise (i.e., $\hat{\mathbf{p}}_j \le \hat{\mathbf{p}}_i$). 

The targe ranking is when all positive pixels are ranked above (have higher confidence than) negative pixels, in which case $\ell^*_{\text{S-Rank}} = 0$.


\noindent\textbf{(2) Spatial Sort Loss} ($\mathcal{L}_{\text{S-Sort}}$) sorts positively labeled pixels with respect to their labels and hence, aligns (smoothed) labels with confidences. 
Following RS Loss \cite{Oksuz_2021_ICCV}, we define our sorting loss as:
\begin{equation}\footnotesize
\mathcal{L}_{\text{S-Sort}} = \frac{1}{|\mathcal{P}|} \sum_{i \in \mathcal{P}} \left( \ell_{\text{S-Sort}}(i) - \ell^*_{\text{S-Sort}}(i) \right),
\end{equation}
where $\ell_{\text{S-Sort}}(i)$ and $\ell^*_{\text{S-Sort}}(i)$ are the current sorting error and target sorting error, respectively. We define the sorting error $\ell_{\text{S-Sort}}(i)$ as the average ``labels'' ($\mathbf{p}$) of pixels with higher ``confidence'' ($\hat{\mathbf{p}}$) than pixel $i$ (i.e., $H(x_{ij}) = 1$):
\begin{equation}\footnotesize
\ell_{\text{S-Sort}}(i) = \frac{1}{\text{rank}^+(i)} \sum_{j \in \mathcal{P}} H(x_{ij})(1 - \mathbf{p}_j),
\end{equation}
where $\mathbf{p}_j \in [0,1]$ is the label of pixel $j$, and $\text{rank}^+(i)=\sum_{j\in\mathbf{P}} H(x_{ij})$ is the number of positives with higher confidence than pixel $i$.

The target sorting error $\ell^*_{\text{S-Sort}}(i)$ is defined as the average label $(1 - \mathbf{p}_j)$ over each positive pixel $j \in \mathcal{P}$ with higher confidence ($H(x_{ij}) = 1$) and higher label ($\mathbf{p}_j \ge \mathbf{p}_i$), calculated as:
\begin{equation}\footnotesize
\ell^*_{\text{S-Sort}}(i) = \frac{\sum_{j \in \mathcal{P}} H(x_{ij}) [\mathbf{p}_j \ge \mathbf{p}_i] (1 - \mathbf{p}_j)}{\sum_{j \in \mathcal{P}} H(x_{ij}) [\mathbf{p}_j \ge \mathbf{p}_i]},
\end{equation}
where $[{P}]$, the Iverson Bracket, is 1 if $P$ is True and 0 otherwise.

\subsection{Instance Sort Loss}

Instance-Sort Loss, $\mathcal{L}_{\text{S-Loss}}$,  aims to improve the alignment between keypoint confidences and their localization qualities. For a certain keypoint type, we define it as:
\begin{equation}\footnotesize
\mathcal{L}_{\text{I-Sort}} = \frac{1}{|\mathcal{I}|} \sum_{i \in \mathcal{I}} \left( \ell_{\text{I-Sort}}(i) - \ell^*_{\text{I-Sort}}(i) \right),
\end{equation}
where $\ell_{\text{I-Sort}}(i)$ and $\ell^*_{\text{I-Sort}}(i)$ are the current instance sorting error and target instance sorting error, respectively. $i$ refers to  a specific person instance in the batch $\mathcal{I}$. Instance sorting error relies  on the localization quality of the predicted pixel within that heatmap  (i.e. the pixel with highest confidence), which is measured by the Keypoint-Similarity score: 
\begin{equation}\footnotesize
\text{KS}(i) = \exp \left( \frac{-d^2}{2s^2 k^2} \right),
\end{equation}

where  $d$ is the distance of the highest-confidence pixel to the ground truth joint location, $s$ is the instance area, and $k$ is the keypoint fall-off value defined (by the COCO dataset \cite{lin2015microsoftcococommonobjects}) for this specific keypoint type. We define the instance  sorting error $\ell_{\text{I-Sort}}(i)$ as the average ``localization quality''  of instances with higher ``confidence'' ($\hat{\mathbf{p}}$) than instance $i$ (i.e., $H(x_{ij}) = 1$): 

\begin{equation}\footnotesize
    \ell_{\text{I-Sort}}(i) = \frac{1}{\text{rank}(i)} \sum_{j\in \mathcal{I}} H(x_{ij}) (1-\text{KS}(j)).
\end{equation}


The target instance sorting error is defined as the average Keypoint Similarity score $(1 - {\text{KS}(j)})$ over each instance $j \in \mathcal{I}$ with higher confidence ($H(x_{ij}) = 1$) and higher KS (${\text{KS}(j)} \ge {\text{KS}(i)}$):
%
\begin{equation}\footnotesize
\ell^*_{\text{I-Sort}}(i) = \frac{\sum_{j \in \mathcal{I}} H(x_{ij}) [{\text{KS}(j)} \ge {\text{KS}(i)}] (1 - {\text{KS}(j)})}{\sum_{j \in \mathcal{I}} H(x_{ij}) [{\text{KS}(j)} \ge {\text{KS}(i)}]},
\end{equation}
%
%
where $[\cdot]$ is the Iverson Bracket.

\textbf{Comparison with Spatial Sort Loss.} Note that the formulation of the Instance Sort Loss is very similar to the formulation of Spatial Sort Loss. The difference is that, for Instance Sort loss, the targets are keypoint similarity scores calculated between the ground truth coordinates and predicted coordinates. Moreover, Instance Sort Loss takes a single pixel value for every keypoint (or heatmap) while Spatial Sort Loss takes multiple pixels as positives for each keypoint.

\subsection{Overall Loss Function and its Optimization}

The overall loss function is a weighted sum of the Spatial Rank, Spatial Sort and Instance Sort losses:
\begin{equation}\footnotesize
\mathcal{L}_{\text{Total}} =  \mathcal{L}_{\text{S-Rank}} + \lambda_1 \mathcal{L}_{\text{S-Sort}} + \lambda_2 \mathcal{L}_{\text{I-Sort}}.
\end{equation}

\noindent{\textbf{Optimization.}} Note that the three introduced loss functions ($\mathcal{L}_{\text{S-Rank}}$, $\mathcal{L}_{\text{S-Sort}}$ and $\mathcal{L}_{\text{I-Sort}}$) include the step function $H(\cdot)$, whose derivative is either zero or undefined. In practice, $H(\cdot)$ is replaced by an approximation with a defined, non-zero gradient in a certain interval \cite{chen2020ap}. Furthermore, despite this approximation, the gradient through $H(\cdot)$ is calculated using an error-driven update mechanism \cite{chen2020ap,Oksuz_2021_ICCV}.
See the supplementary material for more details on optimization of  ranking-based loss functions.

\subsection{How Our Methods Address the Three Problems (P1-P3)}
\label{sect:how_we_address_problems} 

Here we discuss how our methods address the three problems (\textbf{P1-P3}) highlighted in Introduction:

\noindent{\textbf{(P1) Addressing loss not measuring localization quality.} We address this problem with the help of all three losses  ($\mathcal{L}_{\text{S-Rank}},  \mathcal{L}_{\text{S-Sort}}, \mathcal{L}_{\text{I-Sort}}$), which aim to accurately predict positive pixels and their confidences, as well as to align positive pixel confidences with their keypoint similarity scores functioning as a measure of localization quality.

\noindent{\textbf{(P2) Addressing positive-negative imbalance.} Previous work \cite{oksuz2020ranking} showed that a ranking-based loss as our $\mathcal{L}_{\text{S-Rank}}$, by definition, provides balance between the gradients of positive and negative pixels, unaffected by the ratio $\mathcal{N}/\mathcal{P}$ or its effects. 

\noindent{\textbf{(P3) Addressing discrepancy with the evaluation measure.}  Instance Sort Loss aligns the ordering of estimated confidences (heatmap maximums) and their respective keypoint similarity scores. As pointed out by prior work \cite{pmlr-v235-gu24a},  AP depends on the predicted confidences  being rank-consistent with the object keypoint similarity (OKS) scores.  Aligning these two quantities would increase AP. 
We claim that optimizing the alignment between keypoint confidences and their respective localization qualities (Keypoint Similarity) translates to optimizing the alignment between instance confidences and instance localization (OKS) (See the suppl. mat. for a proof).

\section{Experiments}

\textbf{Datasets and Setup.} We evaluate our method on three widely used benchmarks: COCO \cite{lin2015microsoftcococommonobjects}, CrowdPose \cite{Li_2019_CVPR}, and MPII \cite{Andriluka_2014_CVPR}. We apply our ranking-based losses to both 1D classification-based methods (SimCC \cite{li2022simcc}) and 2D heatmap-based methods (ViTPose \cite{NEURIPS2022_fbb10d31}, HRNet-32 \cite{wang2020deep}). On COCO, we evaluate both ViTPose and SimCC with various backbones. For CrowdPose and MPII, we use HRNet-32 as a representative 2D baseline.



\textbf{Performance Metrics.} For COCO and CrowdPose, we report mean Average Precision (mAP) and mean Average Recall (mAR). For MPII, we use Percentage of Correct Keypoints (PCK). Additionally, we compute Spearman's rank correlation coefficient to evaluate the alignment between predicted confidence scores and localization accuracy.



\textbf{Implementation Details.} We follow the default training protocols of the original models unless otherwise stated. All models are trained for 210 epochs with an input resolution of 256\texttimes192. For ViTPose, we use a learning rate of 2e-4 for ViTPose-B and 1e-4 for ViTPose-H, with weight decay of 1e-3. SimCC and HRNet-32 models follow hyperparameter settings from MMPose, except that we tune only the loss-related hyperparameters (delta and loss coefficients). Heatmap resolution is set to 64\texttimes48 for 2D methods and 256\texttimes2/192\texttimes2 for 1D methods.



\begin{table*}
    \centering
    \label{tab:coco_results}
    \footnotesize
    \begin{tabular}{lllc ccc}
        \toprule
        \multirow{1}{*}{Method} & \multirow{1}{*}{Backbone} & \multirow{1}{*}{Loss} & \multirow{1}{*}{mAP} & \multirow{1}{*}{AP$^{50}$} & \multirow{1}{*}{AP$^{75}$} & \multirow{1}{*}{AR} \\
        \midrule
        \multicolumn{7}{l}{\textbf{Regression-based Methods}} \\
        \midrule
        Poseur \cite{Mao_2022_ECCV} & HRFormer-B & MLE & 78.9 & 92.0 & 85.7 & - \\
        \multirow{3}{*}{PCT \cite{Geng_2023_CVPR}}  
            & \multirow{3}{*}{\shortstack[l]{Swin-B\\Swin-L\\Swin-H}} 
            & CE + L1 & 77.7 & 91.2 & 84.7 & - \\
         &  & CE + L1 & 78.3 & 91.4 & 85.3 & - \\
         &  & CE + L1 & 79.3 & 91.5 & 85.9 & - \\
        \midrule
        \multicolumn{7}{l}{\textbf{1D Classification-based Methods}} \\
        \midrule
        \multirow{4}{*}{SimCC \cite{li2022simcc}} 
            & \multirow{2}{*}{ResNet-50} & KL Div. & 72.1 & 89.7 & 79.8 & 78.1 \\
         &  & \textbf{Spatial RS (Ours)} & \textbf{73.6} & \textbf{90.2} & \textbf{80.6} & \textbf{78.8} \\ 
         & \multirow{2}{*}{HRNet-48} & KL Div. & 75.9 & - & - & 81.2 \\
         &  & \textbf{\shortstack{Spatial RS +\\Instance Sort Loss (Ours)}} & \textbf{76.6} & \textbf{90.8} & \textbf{83.4} & \textbf{81.5} \\
        \midrule
        \multicolumn{7}{l}{\textbf{2D Heatmap-based Methods}} \\
        \midrule
        \multirow{4}{*}{ViTPose \cite{NEURIPS2022_fbb10d31}} 
            & \multirow{2}{*}{ViT-B} & MSE & 75.8 & 90.7 & 83.2 & 81.1 \\
         &  & \textbf{\shortstack{Spatial RS +\\Instance Sort Loss (Ours)}} & \textbf{76.6} & \textbf{90.9} & \textbf{83.4} & \textbf{81.7} \\
         & \multirow{2}{*}{ViT-H} & MSE & 79.1 & 91.7 & 85.7 & 84.1 \\
         &  & \textbf{\shortstack{Spatial RS +\\Instance Sort Loss (Ours)}} & \textbf{79.9} & \textbf{92.0} & \textbf{86.4} & \textbf{84.7} \\
        \bottomrule
    \end{tabular}
    \caption{Comparison with the state of the art on COCO-val set.  Input size: 256x192.}
\end{table*}

\subsection{Experimental Results}


\textbf{COCO Results.} Table~\ref{tab:coco_results} summarizes our results on the COCO \cite{lin2015microsoftcococommonobjects} validation set. We compare our ranking-based losses (Spatial-RS + Instance-Sort) against baseline losses across multiple architectures. Our losses improve ViTPose-B and ViTPose-H by 0.8 AP each. For SimCC, we observe gains of 1.5 AP with ResNet-50 and 0.7 AP with HRNet-48.





\begin{table*}
    \centering
    \label{tab:cp_results}
    \footnotesize
    \begin{tabular}{lccccccc}
        \toprule
        Loss & mAP & $\text{AP}^{50}$ & $\text{AP}^{75}$ & $\text{AP}^{E}$ & $\text{AP}^{M}$ & $\text{AP}^{H}$ & AR \\ 
        \midrule
         MSE & 67.8 & 82.4 & 73.6 & 77.1 & 69.0 & 55.9 & 76.8 \\ 
         Spatial Rank & 67.2 & 82.4 & 72.7 & 76.7 & 68.5 & 55.0 & 76.6 \\ 
         Spatial Rank \& Sort & 68.6 & 83.8 & 74.0 & 77.8 & 70.0 & 56.5 & 76.7 \\ 
         Spatial RS + Instance-Sort & \textbf{68.9} & \textbf{84.3} & \textbf{74.3} & \textbf{78.0} & \textbf{70.1} & \textbf{57.4} & \textbf{76.9} \\ 
        \bottomrule
    \end{tabular}
    \caption{Results on CrowdPose dataset using HRNet \cite{wang2020deep} with the HRNetV1-W32 backbone.}
\end{table*}

\textbf{CrowdPose Results.} On CrowdPose \cite{Li_2019_CVPR}, we train HRNet-32 with our proposed losses. As shown in Table~\ref{tab:cp_results}, our method improves AP by 1.1 over the MSE baseline.



\begin{table} 
    \centering
    \label{tab:mpii_results}
    \footnotesize
    \begin{tabular}{lcc}
        \toprule
         Loss & PCK & PCK@0.1 \\ \hline\midrule
         MSE & 90.0 & 33.4 \\ 
         Spatial Rank & 90.6 & 31.2 \\ 
         Spatial Rank \& Sort & \textbf{90.6} & \textbf{34.0} \\ 
        \bottomrule
    \end{tabular}
    \caption{Results on MPII dataset using HRNet \cite{wang2020deep}  with the HRNetV1-W32 backbone.}
\end{table}

\textbf{MPII Results.} On MPII \cite{Andriluka_2014_CVPR}, our method improves HRNet-32 by 0.6 PCK compared to the baseline. Results are summarized in Table~\ref{tab:mpii_results}.





\begin{table} 
    \centering
    \label{tab:corr_results}
    \footnotesize
    \begin{tabular}{lcc}
        \toprule
         Loss &  Spearman Corr. & mAP \\
        \midrule
         MSE &  49.9 & 79.1 \\ 
         Spatial RS  & 55.5 & 79.5 \\ 
        Spatial RS + Instance-Sort &  \textbf{66.5} & \textbf{79.9} \\
        \bottomrule

    \end{tabular}
    \caption{Instance wise correlation on COCO using ViTPose \cite{NEURIPS2022_fbb10d31} with the ViTPose-H backbone.}
\end{table}

\textbf{Correlation between localization quality and confidence} We measure how our losses affect the correlation between localization quality and confidence scores. Previous work \cite{Oksuz_2021_ICCV,kahraman2023correlation} has shown that high correlation between localization qualities and confidences scores are beneficial for the evaluation metric mAP and NMS post processing. In Table~\ref{tab:corr_results} we report correlation coefficients between the localization quality and confidence scores for the ViTPose-H model trained on the COCO dataset. Our results on Table~\ref{tab:corr_results} show that applying only Spatial RS Loss improves Spearman's Ranking Correlation considerably. Applying Instance-Sort Loss on top of Spatial RS Loss also improves the results, resulting in an \textbf{16.6\%} increase in  Spearman's Ranking Correlation Coefficient.

\begin{table*}
    \centering
    \label{tab:comp_ablation}
    \footnotesize
    \begin{tabular}{lllc ccc}
        \toprule
        \multirow{1}{*}{Method} & \multirow{1}{*}{Backbone} & \multirow{1}{*}{Loss Components} & \multirow{1}{*}{mAP} & \multirow{1}{*}{AP$^{50}$} & \multirow{1}{*}{AP$^{75}$} & \multirow{1}{*}{AR} \\
        \midrule
        \multicolumn{7}{l}{\textbf{1D Classification-based Methods}} \\
        \midrule
        \multirow{4}{*}{SimCC \cite{li2022simcc}} & \multirow{4}{*}{ResNet-50} & KL Div. & 72.1 & 89.7 & 79.8 & 78.1 \\
         & & Spatial Rank & 73.0 & 89.8 & 80.2 & 78.4 \\
         & & Spatial Rank \& Sort & \textbf{73.6} & 90.2 & \textbf{80.6} & \textbf{78.8} \\
         & & Spatial RS + Instance-Sort & 73.2 & \textbf{90.4} & 80.5 & 78.6 \\
        \cmidrule(lr){1-7}
        \multirow{4}{*}{SimCC \cite{li2022simcc}} & \multirow{4}{*}{HRNet-48} & KL Div. & 75.9 & - & - & 81.2 \\
         & & Spatial Rank & 76.4 & 90.6 & 82.9 & 81.4 \\
         & & Spatial Rank \& Sort & 76.5 & \textbf{90.8} & 83.2 & 81.5 \\
         & & Spatial RS + Instance-Sort & \textbf{76.6} & \textbf{90.8} & \textbf{83.4} & \textbf{81.5} \\
        \midrule
        \multicolumn{7}{l}{\textbf{2D Heatmap-based Methods}} \\
        \midrule
        \multirow{4}{*}{ViTPose \cite{NEURIPS2022_fbb10d31}} & \multirow{4}{*}{ViT-B} & MSE & 75.8 & 90.7 & 83.2 & 81.1 \\
         & & Spatial Rank & 75.9 & 90.7 & 83.0 & 81.1 \\
         & & Spatial Rank \& Sort & 76.4 & 90.5 & 83.1 & \textbf{81.8} \\
         & & Spatial RS + Instance-Sort & \textbf{76.6} & \textbf{90.9} & \textbf{83.4} & 81.7 \\
        \cmidrule(lr){1-7}
        \multirow{4}{*}{ViTPose \cite{NEURIPS2022_fbb10d31}} & \multirow{4}{*}{ViT-H} & MSE & 79.1 & 91.7 & 85.7 & 84.1 \\
         & & Spatial Rank & 79.1 & 91.6 & 85.7 & 84.0 \\
         & & Spatial RS & 79.5 & 91.4 & 85.9 & 84.6 \\
         & & Spatial RS + Instance-Sort & \textbf{79.9} & \textbf{92.0} & \textbf{86.4} & \textbf{84.7} \\
        \bottomrule
    \end{tabular}
    \caption{Ablation study on loss components (COCO-val results).}
\end{table*}

\subsection{Ablation Study}

\textbf{Impact of Individual Loss Components.}
Our proposed objective comprises three components: Spatial Rank Loss, Spatial Sort Loss, and Keypoint Similarity Sort (Instance-Sort) Loss. We conduct an ablation study to isolate the contribution of each component to the final model performance. Results are presented in Table~\ref{tab:comp_ablation}.

Applying Spatial Rank Loss alone leads to consistent improvements in mAP and mAR across most models and backbones. The only exception is observed on CrowdPose, where Spatial Rank Loss underperforms compared to the baseline MSE loss. We attribute this to the high prevalence of human-to-human occlusions in CrowdPose, which may reduce the effectiveness of relative ranking supervision when used in isolation.

Adding Spatial Sort Loss further improves performance in all cases, suggesting that explicitly supervising the relative ordering of keypoints reinforces the signal provided by Spatial Rank Loss. Finally, incorporating Instance-Sort Loss yields additional gains in nearly all settings, with the exception of SimCC with ResNet-50, where its contribution is marginally negative. These results underscore the complementary nature of the three loss components.

\begin{table} 
    \centering
    \label{tab:coeff_ablation_merged}
    \footnotesize
    \setlength{\tabcolsep}{6pt} 
    \begin{tabular}{cccc ccccc}
        \toprule
        \multicolumn{4}{c}{\textbf{Spatial Sort Loss}} &
        & \multicolumn{4}{c}{\textbf{Instance Sort Loss}} \\
        \cmidrule(r){1-4} \cmidrule(l){6-9}
        Delta & Coeff. & AP & AR &
        & Delta & Coeff. & AP & AR \\
        \cmidrule(lr){1-4} \cmidrule(lr){6-9} 
        \multirow{3}{*}{1.0} & 0.5 & 76.2 & 81.4 & & \multirow{3}{*}{2.0} & 0.25 & 76.3 & 81.6 \\
                             & 1.0 & 76.2 & 81.4 & &                        & 0.50 & 76.5 & \textbf{81.7} \\
                             & 2.0 & 76.2 & 81.5 & &                        & 1.00 & 76.4 & 81.6 \\
        \cmidrule(lr){1-4} \cmidrule(lr){6-9} 
        \multirow{3}{*}{1.5} & 0.5 & 76.2 & 81.5 & & \multirow{3}{*}{3.0} & 0.25 & \textbf{76.6} & \textbf{81.7} \\
                             & 1.0 & 76.1 & 81.4 & &                        & 0.50 & 76.5 & \textbf{81.7} \\
                             & 2.0 & \textbf{76.4} & \textbf{81.8} & &      & 1.00 & 76.5 & \textbf{81.7} \\
        \bottomrule
    \end{tabular}
    \caption{Ablation analysis of deltas and coefficients for ViTPose-B on COCO. The Spatial Sort Loss is used together with the Spatial Rank Loss (delta = 0.4, coefficient = 1.0). Additionally, the Instance Sort Loss is combined with the Spatial Rank Loss (delta = 0.4, coefficient = 0.1) and the Spatial Sort Loss (delta = 1.5, coefficient = 2.0). }
\end{table}

\textbf{Effect of Delta and Coefficients.}
We also investigate the impact of varying the delta ($\delta$) values and weighting coefficients associated with each loss term. Unlike prior work in ranking-based objectives \cite{Oksuz_2021_ICCV, Cetinkaya_2024_CVPR, yavuz2024bucketed}, we decouple the deltas used in the rank and sort losses, allowing for more flexible tuning. Table~\ref{tab:coeff_ablation_merged} reports the performance under different hyperparameter configurations.

\subsection{Efficiency Analysis}

See the supplementary material for an analysis of the efficiency of the proposed loss functions.

\section{Conclusion}
In this paper, we propose ranking-based loss functions for human pose estimation to address key limitations of heatmap representations and the commonly used Mean Squared Error (MSE) loss. MSE penalizes all pixel deviations equally, without explicitly emphasizing sharp and accurate localization of joint peaks. Moreover, due to the significant imbalance between positive and negative pixels, label smoothing is typically required. Another critical issue is the misalignment between the training objective (MSE loss) and the evaluation metric (mAP). Our ranking-based losses mitigate the imbalance and are better aligned with the evaluation metric. We demonstrate the effectiveness of our approach on both one-dimensional and two-dimensional heatmaps, using three different backbones and three benchmark datasets. Our method surpasses the previous state of the art on the COCO-val set, achieving 79.9 AP with ViTPose-H, a vision transformer model. We also improve SimCC ResNet-50, a coordinate classification-based method, by 1.5 AP, reaching 73.6 AP on COCO-val.

\paragraph{Limitations} Although our proposed losses yield better average precision and its per iteration overhead over the traditional MSE looks minimal ($1.52$ seconds per iteration vs. $1.427$ seconds), this overhead might add up when a large model is trained. For example, full training of ViT-H model takes $4.33$ days when using our losses, and $4.06$ days when using MSE loss. In addition, our losses use $\sim 2\%$ more GPU memory than the MSE loss. Additionally, since our loss function includes three delta and three coefficient parameters, hyperparameter optimization becomes more challenging.

\section{Acknowledgments}
We gratefully acknowledge the computational resources provided by METU-ROMER, Center for Robotics and Artificial Intelligence, Middle East Technical University, and TUBITAK ULAKBIM High Performance and Grid Computing Center (TRUBA). Emre Akbas and Sinan Kalkan were supported by the Middle East Technical University Scientific Research Projects (BAP) program through project ADEP-312-2024-11485, titled ``New Techniques in Visual Recognition.''



\appendix
\section{Optimization of Ranking Based Losses}

The losses $\mathcal{L}_{\text{S-Rank}}$, $\mathcal{L}_{\text{S-Sort}}$ and $\mathcal{L}_{\text{I-Sort}}$ are defined via discrete counts of pairwise relationships between pixel scores, involving the Heaviside step function $H(\mathbf{p}_j - \mathbf{p}_i)$, which is either non‐differentiable or has zero (uninformative) derivative.  Consequently, the exact derivative
\begin{equation}
\frac{\partial \mathcal{L}_o}{\partial p_i},
\qquad o \in \{\text{S-Rank},\,\text{S-Sort},\,\text{I-Sort}\},
\end{equation}
vanishes on almost all inputs, preventing direct backpropagation.

To address this, we adopt the error‐driven update rule of Chen et al. \cite{chen2020ap}, originally developed for AP‐Loss, which replaces the true (zero‐almost‐everywhere) gradient with a sum of pairwise correction terms.  Specifically, we approximate
\begin{equation}
\frac{\partial \mathcal{L}_o}{\partial p_i}
\;\approx\;
\sum_{j}{L}^o_{j i}\,t^o_{j i}
\;-\;
\sum_{j}{L}^o_{i j}\,t^o_{i j},
\label{eq:error_driven}
\end{equation}
where
\begin{itemize}
  \item ${L}^o_{ij}$ is called the primary term quantifying \emph{pairwise error contribution} between pixels $i$ and $j$ under loss $o$.  For ranking loss,
  \begin{equation}
    {L}^{\mathrm{S-Rank}}_{ij}
    = 
    \frac{H(p_j - p_i)}
         {\displaystyle\sum_{k\in P\cup N} H(p_k - p_i)},
  \end{equation}
  and analogously for $\mathcal{L}_{\mathrm{S-Sort}}$ and $\mathcal{L}_{\mathrm{I-Sort}}$.
  \item $t^o_{ij}\in\{0,1\}$ is a \emph{pair indicator} selecting only the relevant comparisons:
  \begin{equation}
    t^{\mathrm{Rank}}_{ij} =
    \begin{cases}
      1, & y_i=1 \land y_j=0,\\
      0, & \text{otherwise},
    \end{cases}
    \quad
    t^{\mathrm{Sort}}_{ij} =
    \begin{cases}
      1, & y_i=1 \land y_j=1,\\
      0, & \text{otherwise}.
    \end{cases}
  \end{equation}
\end{itemize}

Despite the original losses being defined via non‐differentiable counts, this approximation yields a usable gradient estimate for end‐to‐end training.  For full derivation, see Chen et al. \cite{chen2020ap}.

\section{Optimizing the alignment between OKS and instance confidence scores}

We claim that improving the alignment between keypoint confidences and their localization quality (Keypoint Similarity) leads to better alignment between instance confidence and instance localization (OKS). In Theorem 1, we show that when the covariance between keypoint confidences and their localization quality increases, the covariance between instance confidence and instance localization quality (OKS) also increases. Prior work \cite{pmlr-v235-gu24a} has shown that mean Average Precision (mAP) depends on the consistency of the ordering between instance confidences and instance localization quality (OKS).

\bigskip




\begin{theorem}
Let \( L_k \in \mathbb{R} \) denote the localization quality for keypoint \( k \),  
and let \( C_k \in \mathbb{R} \) denote the model's confidence score for keypoint \( k \). Assume independent keypoints, i.e., \( \text{Cov}(L_{\cdot,j}, C_{\cdot,m}) = 0 \) for \( j \ne m \).  
If \( \sum_{k=1}^K \text{Cov}(L_k, C_k) \) increases, then \( \text{Cov}(\text{OKS}, \text{Conf}) \) increases,  
where \( \text{OKS} = \sum_{k=1}^K L_k \) and \( \text{Conf} = \sum_{k=1}^K C_k \).
\end{theorem}

\bigskip
\begin{proof}
\textbf{Proof.} The covariance between OKS and Conf over \(n\) instances is given by:
\begin{equation}
\text{Cov}(\text{OKS}, \text{Conf}) = \frac{1}{n} \sum_{i=1}^n (\text{OKS}_i - \mathbb{E}[\text{OKS}]) (\text{Conf}_i - \mathbb{E}[\text{Conf}])
\end{equation}

Expanding \(\text{OKS}_i\) and \(\text{Conf}_i\):
\begin{equation}
\text{Cov}(\text{OKS}, \text{Conf}) = \frac{1}{n} \sum_{i=1}^n \left( \frac{1}{K} \sum_{j=1}^K L_{i,j} - \mathbb{E}[\text{OKS}] \right) \left( \frac{1}{K} \sum_{j=1}^K C_{i,j} - \mathbb{E}[\text{Conf}] \right)
\end{equation}

Using the linearity of covariance and expectation:
\begin{equation}
\text{Cov}(\text{OKS}, \text{Conf}) 
= \frac{1}{K^2} \sum_{j=1}^K \sum_{m=1}^K \frac{1}{n} \sum_{i=1}^n \left(L_{i,j} - \mathbb{E}[L_{\cdot,j}]\right) \left(C_{i,m} - \mathbb{E}[C_{\cdot,m}]\right).
\end{equation}

This simplifies to:
\begin{equation}
\text{Cov}(\text{OKS}, \text{Conf}) = \frac{1}{K^2} \sum_{j=1}^K \sum_{m=1}^K \text{Cov}(L_{\cdot,j}, C_{\cdot,m})
\end{equation}

Assuming independent keypoints, i.e., \(\text{Cov}(L_{\cdot,j}, C_{\cdot,m}) = 0\) for \(j \ne m\), we obtain:
\begin{equation}
\text{Cov}(\text{OKS}, \text{Conf}) = \frac{1}{K} \sum_{j=1}^K \text{Cov}(L_{\cdot,j}, C_{\cdot,j})
\end{equation}

Thus, increasing any \(\text{Cov}(L_k, C_k)\) increases \(\text{Cov}(\text{OKS}, \text{Conf})\). \hfill \(\blacksquare\)
\end{proof}

\section{Ablation: Effect of Confidence Scaling}

Ranking‐based losses optimize  the relative ordering of predictions, and as a result the raw logits often span a much larger dynamic range than in standard detection models. Since many downstream applications expect confidence scores to lie in a fixed interval for interpretability and stability, we apply a simple per‐keypoint min–max normalization using the observed minima and maxima from the COCO \texttt{train2017} \cite{lin2015microsoftcococommonobjects} set. We evaluate the ViT‐H \cite{NEURIPS2022_fbb10d31} model on COCO \texttt{val2017} \cite{lin2015microsoftcococommonobjects} both before and after this scaling. We observe no change in performance—average precision remains at 79.9\% and average recall at 84.7\%—demonstrating that straightforward range‐constrained scaling yields interpretable confidences without sacrificing accuracy.

\section{Efficiency Analysis}

Ranking-based losses are often associated with increased computational overhead, particularly due to the large number of pairwise comparisons required. Prior work \cite{yavuz2024bucketed} addressed this by proposing efficient ranking objectives for object detection, enabling their use with large transformer-based backbones. 
In our setting, the difference in computational cost compared to the baseline is negligible. This is primarily due to the smaller number of negative pairs in pose estimation compared to object detection. To quantify this, we benchmark the per-iteration training time for ViTPose-H \cite{NEURIPS2022_fbb10d31} using standard MSE Loss versus our full loss (Spatial-RS + Instance-Sort). Experiments are conducted on 2 NVIDIA H100 GPUs with a batch size of 64 per GPU. Averaged over 1,000 iterations, MSE Loss takes 1.427 seconds per iteration, while our proposed loss takes 1.520 seconds. This indicates that our method introduces minimal overhead relative to conventional losses. Compared to MSE in the previously described setting, our method uses only $\sim 2\%$ more GPU memory.

\section{Experimental Details}

For the SimCC \cite{li2022simcc} and HRNet \cite{wang2020deep} experiments, we follow the official MMPose \cite{mmpose2020} configurations and use the Adam optimizer. For ViTPose, we also adopt the MMPose \cite{mmpose2020} configuration, using the AdamW optimizer with a weight decay of 1e-3.

We use a batch size of 64 for all experiments conducted on the COCO and CrowdPose \cite{Li_2019_CVPR} datasets. For experiments on MPII \cite{Andriluka_2014_CVPR}, we use a batch size of 16.

A multi-step learning rate schedule is used, with decay steps at epochs 170 and 200, and a decay factor (gamma) of 0.1.

\paragraph{Loss Hyperparameters.} We tune the delta values and weighting coefficients for the \textit{Spatial Rank}, \textit{Spatial Sort}, and \textit{Instance Sort} loss components. The optimal hyperparameter settings used in our experiments are provided in Table~\ref{tab:loss_hyper}.

\begin{table}[ht]
\centering
\label{tab:loss_hyper}

\begin{tabular}{l|l|cc|cc|cc}
\toprule
Method & Backbone & \multicolumn{2}{c|}{S. Rank} & \multicolumn{2}{c|}{S. Sort} & \multicolumn{2}{c}{Ins. Sort} \\
       &          & Delta & Coeff & Delta & Coeff & Delta & Coeff \\
\midrule
ViTPose \cite{NEURIPS2022_fbb10d31} & ViT-B     & 0.4 & 1.0 & 1.5 & 2.0 & 3.0 & 0.25 \\
ViTPose \cite{NEURIPS2022_fbb10d31} & ViT-H     & 0.4 & 1.0 & 1.5 & 2.0 & 3.0 & 1.5 \\
SimCC \cite{li2022simcc}   & ResNet-50 & 0.4 & 1.0 & 0.2 & 0.25 & - & - \\
SimCC \cite{li2022simcc}  & HRNet-W48 & 0.3 & 1.0 & 1.2 & 0.5 & 0.5 & 0.1 \\
\bottomrule
\end{tabular}
\caption{Loss hyperparameters (delta and coefficient) for each method and backbone for the COCO dataset.}
\label{tab:loss_hyperparams}
\end{table}

\section{Computational Resources}

SimCC with HRNet-48 backbone, HRNet-32, and ViTPose-B were trained on two A100 GPUs; SimCC ResNet-50 used a single A100 GPU. ViTPose-H was trained on two H100 GPUs. Training the largest model, ViTPose-H, on COCO took approximately 4.5 days on two H100 GPUs.

\end{document}